# Searth Transformer: A Transformer Architecture Incorporating Earth's Geospheric Physical Priors for Global Mid-Range Weather Forecasting

Tianye Li[1,3], Qi Liu[2,3], Hao Li[4,5], Lei Chen[4,5], Wencong Cheng[6,*], Fei Zheng[2], Xiangao Xia[7,3], Ya Wang[8], Gang Huang[2], Weiwei Wang[9,10], Xuan Tong[11], Ziqing Zu[12,13], Yi Fang[1], Shenming Fu[2], Jiang Jiang[14], Haochen Li[15], Mingxing Li[1,3], Jiangjiang Xia[1,3,*]

[1]Temperate East Asia Research Center on Global Change, Institute of Atmospheric Physics, Chinese Academy of Sciences, Beijing 100029, China
[2]State Key Laboratory of Earth System Numerical Modeling and Application, Institute of Atmospheric Physics, Chinese Academy of Sciences, Beijing 100029, China
[3]University of Chinese Academy of Sciences, Beijing 100049, China
[4]Artificial Intelligence Innovation and Incubation Institute, Fudan University, Shanghai, 200433, China.
[5]Shanghai Academy of Artificial Intelligence for Science, Shanghai, 200232, China
[6]Beijing Aviation Meteorological Institute, Beijing 100085, China
[7]Laboratory of Middle Atmosphere and Global Environment Observation (LAGEO), Institute of Atmospheric Physics, Chinese Academy of Sciences, Beijing 100029, China
[8]Earth System Numerical Simulation Science Center, Institute of Atmospheric Physics, Chinese Academy of Sciences, Beijing 100029, China
[9]Meteorological Bureau of Shenzhen Municipality, Shenzhen 518040, China
[10]Shenzhen Key Laboratory of Severe Weather in South China, Shenzhen 518040, China
[11]Jiangsu Key Laboratory of Intelligent Weather Forecasting and Applications Based on Big Data, Nanjing University of Information Science and Technology, Nanjing 210044, China.
[12]State Key Laboratory of Satellite Ocean Environment Dynamics, National Marine Environmental Forecasting Center, Beijing 100081, China
[13]Key Laboratory of Marine Hazards Forecasting, Ministry of Natural Resources, Beijing 100081, China
[14]Beijing Meteorological Service Center, Beijing 100089, China
[15]School of Science, Beijing University of Posts and Telecommunications, Beijing 100876, China

**Corresponding authors:** Jiangjiang Xia, xiajj@tea.ac.cn; Wencong Cheng, emailtocheng@sina.com

**Abstract**—Accurate global medium-range weather forecasting is pivotal to Earth system science and serving as a critical public-service application. While modern weather forecasting increasingly employs data-driven AI models, most Transformer-based architectures rely on generic vision-centric designs that overlook the Earth's inherent spherical topology and zonal periodicity. Additionally, the resource-intensive nature of autoregressive training imposes critical bottlenecks, constraining attainable forecast horizons while aggravating error propagation. These limitations not only compromise physical consistency and forecast accuracy but also effectively preclude resource-constrained institutions from leveraging localized datasets to refine global model performance. To address these challenges, this paper proposes the Shifted Earth Transformer (Searth Transformer), a physics-informed transformer architecture designed for global medium-range weather forecasting. Searth Transformer integrates zonal periodicity and meridional boundaries into window-based self-attention, enabling physically consistent global information exchange. To mitigate the computational bottlenecks, we develop the Relay Autoregressive (RAR) fine-tuning strategy, a memory-efficient strategy decoupling GPU memory usage from the forecast length. This enables the model to learn long-range atmospheric evolution and suppress the accumulation of forecast errors within constrained memory footprints. Leveraging these innovations, we present YanTian, a global medium-range weather forecasting model. Despite using about 1/200 of the computational resources required by standard autoregressive fine-tuning methods, YanTian demonstrates superior accuracy over the European Centre for Medium-Range Weather Forecasts (ECMWF)'s high-resolution forecast (HRES) and achieves performance comparable to state-of-the-art AI-based medium-range weather forecasting models at a spatial resolution of 1°. Meanwhile, YanTian achieves a longer skillful forecast lead time for Z500 (10.3 days) than ECMWF HRES (9 days). Beyond weather forecasting, this work establishes a robust algorithmic foundation for the predictive modeling of complex, global-scale geophysical circulation systems, offering new pathways for Earth system science.

## I. INTRODUCTION

Over the past decade, the Transformer architecture has undergone a remarkable transition from natural language processing to computer vision [1], [2], [3] and, more recently, to the natural sciences, demonstrating strong modeling capacity and broad applicability [4], [5], [6]. Crucially, the success of Transformers beyond the language domain has relied not on direct transplantation, but on substantial architectural reinterpretation to accommodate domain-specific data structures. In computer vision, this transition was first realized by the Vision Transformer (ViT) [2], which reformulated images as sequences of patch tokens, enabling the application of global self-attention originally designed for language modeling. However, visual data exhibit inherent spatial locality and multi-scale hierarchies that differ fundamentally from linguistic sequences. To better align with these characteristics, the Swin Transformer [7] introduced hierarchical feature representations and localized, shifted window-based self-attention, substantially reducing the computational burden of global attention while preserving modeling flexibility. Together, ViT and Swin Transformer exemplify how the core Transformer paradigm can be systematically adapted from natural language to vision, striking an effective balance between computational tractability and expressive power. More broadly, this evolution highlights a key principle: the full potential of Transformer architectures emerges only when their design is explicitly aligned with the structural priors of the target domain.

A parallel momentum is currently reshaping the landscape of Earth system modeling [8], where Transformer-based architectures are igniting a paradigm shift in global medium-range weather forecasting. In recent years, the field has witnessed the emergence of numerous state-of-the-art global weather models based on the Swin Transformer architecture—such as Pangu-Weather [5], FuXi [9], FengWu [10], and Aurora[11]. These models demonstrate that at a 0.25° resolution, Transformer-based architectures can rival or even exceed the accuracy of operational numerical weather prediction (NWP) systems, such as the European Centre for Medium-Range Weather Forecasts (ECMWF)'s high-resolution forecast (HRES) [12].

However, the limitations of directly transferring the Swin Transformer from computer vision to Earth system modeling have become increasingly evident and now constitute a key bottleneck for further progress. Specifically, the Swin Transformer is predicated on the assumption that data are situated on a Euclidean planar grid characterized by disconnected boundary conditions. This assumption fundamentally conflicts with the true physical topology of the Earth system, which exhibits zonally periodic continuity and meridional boundaries on a spherical surface. As a result, there is currently a lack of Transformer variants that explicitly encode the zonally periodic connectivity of the Earth system as a physical prior for Earth system modeling. This geometric discrepancy imposes an inherent upper bound on restricts the representation of large-scale circulation patterns and global teleconnections, ultimately impairing forecast skill due to physically inconsistent information exchange across boundaries. Especially as the forecast lead time increases, it exacerbates the rapid accumulation of errors caused by the lack of exchange of zonal information.

Additionally, in data-driven AI-based weather forecasting, global medium-range prediction is typically formulated as an autoregressive problem, where deep learning models repeatedly apply one-step forecasts to advance the atmospheric state over extended lead times [13], [14]. However, since the atmosphere is an inherently chaotic dynamical system, forecast errors tend to accumulate rapidly as the prediction horizon extends [15]. Consequently, many weather forecasting models adopt a two-stage training paradigm: (1) single-step forecasting training followed by (2) accumulated-error suppression training [9], [10], [16], [17]. In the first stage, models are trained on large-scale global reanalysis datasets to acquire the capability of predicting the weather state one step ahead. To mitigate the error accumulation, the second training stage is typically introduced to learn the continuous evolution of weather processes over multiple future steps. However, during this multi-step learning phase, the temporal length of the target atmospheric evolution is rigidly coupled with the GPU memory footprint required for autoregressive rollout, as intermediate model states across successive forecast lead times must be explicitly stored for backpropagation through time [18]. As a result, extending the prediction horizon to capture long-term atmospheric evolution substantially increases computational and memory costs, severely constraining scalability and rendering long-range weather forecasting with high spatial resolution prohibitively expensive [19].

To address these challenges simultaneously, we proposed two key innovations. First, we propose the Shifted Earth Transformer (Searth Transformer), a Transformer architecture for Earth system modeling that explicitly incorporates the physical prior of zonally periodic continuity and meridional boundaries. Specifically, the Searth Transformer integrates zonal periodicity and meridional boundaries into window-based self-attention via an asymmetric shift-and-mask mechanism. Unlike standard architectures, this architecture eliminates the attention masks between the zonal (East-West) boundaries to facilitate seamless periodic connectivity and longitudinal wrap-around information exchange. Conversely, it retains the masks along meridional (North-South) boundaries to respect the physical constraints of the poles, thereby ensuring that self-attention remains strictly aligned with the geophysical topology of the Earth system. Second, we develop the Relay Autoregressive (RAR) fine-tuning strategy, a dedicated paradigm that decouples GPU memory overhead from the forecast length. Specifically, the long-term atmospheric evolution is decomposed into multiple shorter sub-stages, each of which is trained independently and backpropagated sequentially. At the boundary of each stage, gradient detachment is applied such that the final prediction of the sub-stage is detached from the computational graph and used as the input to the next sub-stage. This relay-style learning strategy enables the model to learn long-term weather evolution by progressively inheriting accumulated forecast errors and physical biases, while ensuring that GPU memory usage is determined solely by the length of each sub-stage rather than the full forecast horizon. Building upon these innovations, we developed YanTian (means "forecast the weather" in Chinese [18]), a global medium-range weather forecasting model trained on 1° ERA5 reanalysis data. In the evaluation of forecasts at 1° horizontal resolution, the Yantian model outperforms the ECMWF HRES [20] and achieves accuracy comparable to state-of-the-art AI models.

Our key contributions can be summarized as follows.
1) Propose the Shifted Earth Transformer (Searth Transformer), a Transformer architecture that incorporates zonal periodicity and meridional boundary constraints, enabling seamless longitudinal information exchange while respecting polar physical limits.
2) Develop the Relay Autoregressive (RAR) fine-tuning strategy, which decomposes long-term atmospheric evolution into sequential sub-stages with gradient detachment, allowing efficient long-range fine-tuning without excessive GPU memory usage.
3) A 1°-resolution global medium-range weather forecasting model, YanTian, was built using the Searth Transformer and RAR fine-tuning, demonstrating that improved physical consistency can significantly enhance modeling capability and offset the limitations of coarse resolution and limited computational resources.

## II. RELATED WORKS

*A. Applications of Swin Transformer in Medium-Range Weather Forecasting*

Swin Transformer has been extensively adopted in recent global medium-range weather forecasting models [6], [9], [10], [11]. These models can currently be categorized into 3D attention mechanisms and 2D attention mechanisms according to their attention mechanisms. Models with 3D attention mechanisms treat the global atmosphere as a three-dimensional cube and compute attention within each 3D window. Bi et al. [5] developed PanGu-Weather, which incorporates three-dimensional attention mechanisms and Earth position encoding to explicitly model atmospheric structures. The training of each model takes approximately 16 days on a cluster of 192 NVIDIA Tesla V100 GPUs. Bodnar et al. [11] developed Aurora, the first Earth system foundation model using a 3D Swin-Transformer that demonstrates cross-domain generalization capabilities. During the pretraining phase, 32 A100 GPUs were employed, requiring approximately 2.5 weeks. Although 3D attention mechanisms offer strong modeling capabilities, their cubic computational complexity $O(n^3)$ makes model training significantly more resource-intensive compared to 2D attention mechanisms.

Some global medium-range weather forecasting models directly adopt 2D Swin-Transformers, treating the global atmosphere as a two-dimensional bounded plane. Chen et al [9]. developed FuXi, a cascaded model with 4.5 billion parameters, which demonstrates strong forecasting capabilities. FengWu treats weather forecasting as a multimodal problem, enabling operational medium-range deterministic forecasts to be extended beyond a 10-day lead time [10]. These designs benefit from the scalability and powerful modeling capabilities of the 2D Swin Transformer, achieving high accuracy in medium-range weather forecasts at 0.25° resolution and easily scaling the number of parameters to the billion-level range.

However, whether the Swin-Transformer employs 3D or 2D attention, it neglects the zonally periodic continuity and meridional boundaries inherent to a spherical surface [21], [22]. These physical inconsistencies constrain the modeling capability of the underlying algorithms and additionally increase the training cost, as the model must learn the true atmospheric evolution within an incorrect physical space. Therefore, it is necessary to incorporate geophysical priors of zonally periodic continuity and meridional boundaries into the Swin Transformer, creating a dedicated Transformer architecture specifically designed for modeling the Earth's circulation system.

*B. Strategies for Suppressing Error Accumulation*

Existing approaches for suppressing rapid error accumulation in autoregressive forecasting can be broadly categorized into single-model and multi-model strategies.

Among single-model approaches, GraphCast represents a typical example [16]. It adopts a curriculum learning strategy combined with autoregressive fine-tuning, where the model is initially trained with two autoregressive steps and gradually extended to twelve steps. This method enables the model to learn continuous atmospheric evolution within a limited temporal horizon. Nevertheless, autoregressive fine-tuning necessitates constructing a complete computational graph over the entire forecast trajectory, resulting in GPU memory consumption that scales linearly with the forecast horizon. Due to hardware constraints, GraphCast was only able to learn weather evolution within a three-day window (12 steps), which is insufficient for medium-range forecasting.

Multi-model approaches attempt to address this issue by decomposing long-range forecasting into multiple shorter-range models. Pangu-Weather introduces a hierarchical temporal aggregation strategy by training four separate models [5], each responsible for a different forecast lead time, with the longest interval being 24 hours. Although this design reduces the number of autoregressive steps and partially mitigates error accumulation, it requires multiple models, thereby substantially increasing computational cost. FuXi further extends GraphCast's autoregressive fine-tuning paradigm by proposing a cascaded architecture [9], in which three models are separately fine-tuned for forecast ranges of 0–5 days, 5–10 days, and 10–15 days. While this design effectively extends the forecast horizon, each model remains constrained by GPU memory limitations and can only learn the first three days of atmospheric evolution within its assigned range. So noticeable performance degradation is observed at the transition boundaries between adjacent models. Meanwhile, training three models also need more computational cost.

In summary, there remains an urgent need to develop a transformer architecture that is inherently suitable for modeling Earth system circulation while simultaneously proposing a resource-efficient fine-tuning strategy capable of learning atmospheric evolution over long forecast horizons.

III. METHOD

*A. YanTian Model Architecture Overview*

In this section, we introduce the overall neural network architecture of YanTian, a global medium-range weather forecasting model built upon a physics-constrained Transformer architecture and a compute-efficient fine-tuning strategy, as showed in **Fig. 1** (a). YanTian model is designed for a forecasting interval of 6 hours and leverages weather parameters from two previous time steps, denoted as $(X_{t-6h}, X_t)$, to predict the atmospheric state at the next time step $X_{t+6h}$. Detailed descriptions of the input and output variables are provided in section IV. EXPERIMENTS AND ANALYSIS. After being fed into the model, the input variables propagate through two distinct pathways: a main computational pathway responsible for predicting state differences between successive time steps, and a skip-connection pathway that provides a physical reference state for the final prediction [24].

The main computational pathway follows an Encoder-Core-Decoder paradigm [25].

The input $(X_{t-6h}, X_t)$ is first processed by the Encoder to perform downsampling and compact representation learning. Specifically, an embedding layer implemented as a 3D convolutional neural network is employed to jointly encode temporal and spatial information [26], [23], compressing the input from two time-step weather states at 1° resolution (2×69×180×360) into latent space (768×90×180). The embedded features are then passed through six stacked Searth Transformer blocks with residual connections [28]. Subsequently, a patch-merging module, following the design in Swin Transformer [7], further downsamples the latent representation, resulting in a latent tensor of size 1536 × 45 × 90.

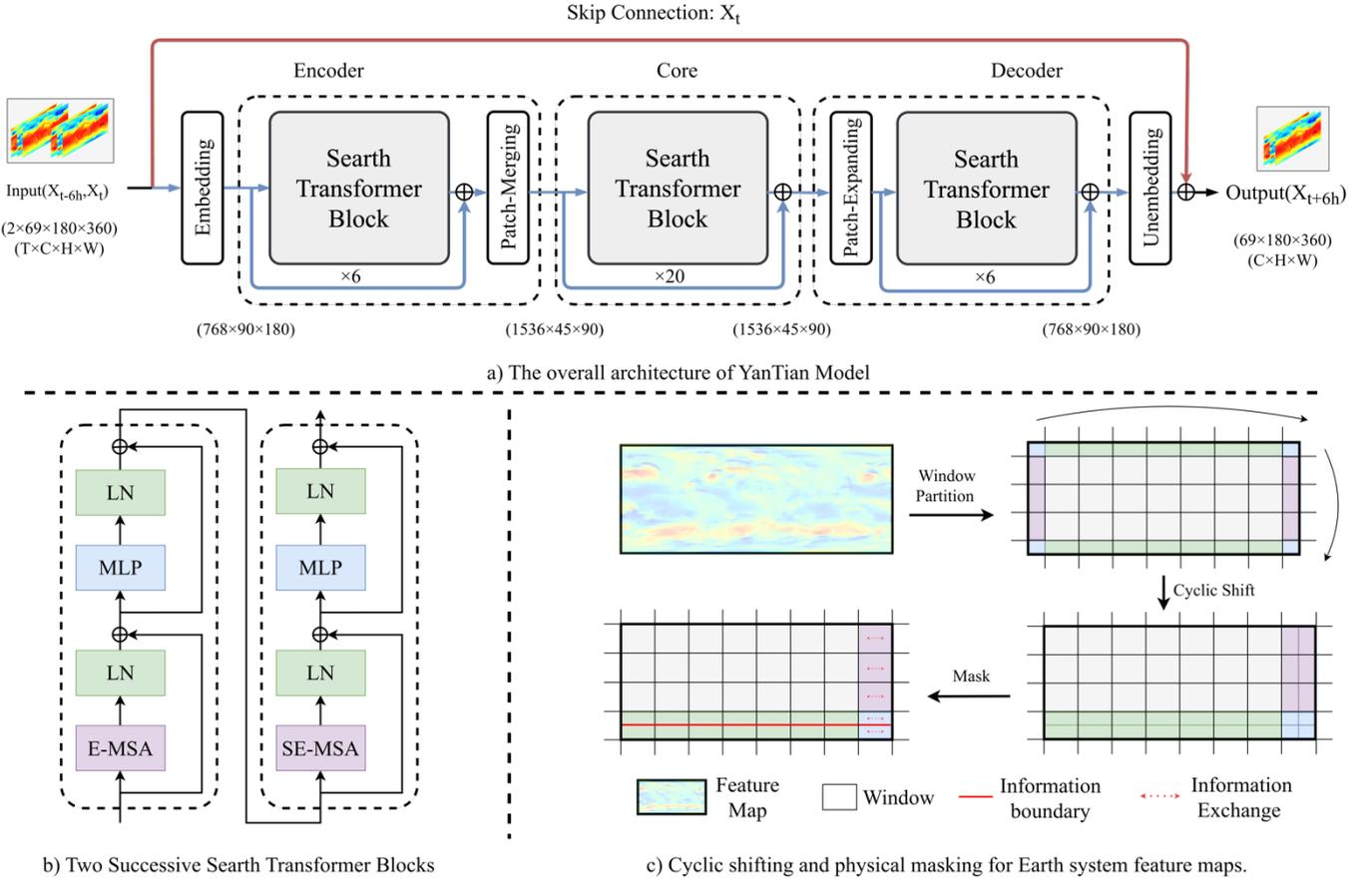

**Fig. 1.** Architectural overview of YanTian model. (a) The hierarchical encoder-core-decoder framework utilizing Shifted Earth Transformer blocks and multiple skip connections. (b) Two successive Shifted Earth Transformer Blocks. E-MSA is algorithmically identical to W-MSA. SE-MSA is an enhanced variant of the SW-MSA module in the Swin Transformer, as illustrated in (c). (c) This modification incorporates the spherical physical characteristics of the Earth system—East-West periodic connectivity and North-South discontinuity—into the cyclic shift and masking operations, where the masks between the east–west boundaries are removed after cyclic shifting to enable cross-boundary information exchange, while the masks at the north–south boundaries are preserved.

The latent tensors are then fed into the Core module, which consists of 20 stacked Searth Transformer blocks with residual connections. This stage focuses on capturing large-scale atmospheric evolution patterns.

Afterward, the Decoder progressively reconstructs higher-resolution representations. The latent variables are first upsampled to a spatial size of 90 × 180 with 768 feature channels via a patch-expanding operation, followed by six additional Searth Transformer blocks with residual connections. Finally, an unembedding layer decodes the latent variables into the predicted weather state tendency (69×180×360). The unembedding layer is composed of a 2D transposed convolutional network and a fully connected layer [27].

Built upon the physics-informed Searth Transformer, the Encoder–Core–Decoder architecture achieves a favorable balance between computational efficiency and physical realism, enabling accurate prediction of atmospheric state variations.

In parallel, the skip-connection pathway directly extracts $X_t$ as a baseline state and forwards it to the end of the network, where it is added to the predicted state differences between successive time steps produced by the main pathway. This design allows the model to focus on forecasting the incremental atmospheric evolution rather than reconstructing the full weather state, thereby improving predictive accuracy [24].

Overall, the YanTian architecture is characterized by two key design features. First, it introduces the Searth Transformer, a Swin Transformer variant tailored for Earth system modeling that explicitly enhances zonal information flow. Second, extensive use of skip connections throughout the network stabilizes training and further enhances predictive skill. The total number of parameters in YanTian reaches 600 million.

*B. Shifted Earth Transformer*

To incorporate the intrinsic physical topology of zonal periodicity and meridional boundary constraints into the Swin Transformer, we propose the Searth Transformer. The core design is to reconcile the window-based self-attention with the physically continuous zonal (east–west) circulation and the inherently non-periodic meridional (north–south) structure of the atmosphere.

As illustrated in **Fig. 1** (b), the Searth Transformer adopts hierarchical blocks consisting of two successive transformer blocks, following a design similar to that of the Swin Transformer [7]. The first block employs Earth-aware Multi-head Self-Attention (E-MSA), which is algorithmically equivalent to the standard window-based multi-head self-attention (W-MSA). This block focuses on efficient local feature aggregation within non-overlapping windows while preserving computational scalability for high-resolution global fields. Following a residual connection, layer normalization, and a feed-forward MLP, the representation is forwarded to the second block.

The second block introduces the key novelty of this work, named Shifted Earth Multi-head Self-Attention (SE-MSA). As showed in **Fig. 1** (c), SE-MSA performs cyclic shifts along both the zonal and meridional directions to enable cross-window information exchange. However, unlike conventional shifted-window attention, the masking strategy is explicitly redesigned according to physical priors of the Earth system: the east-west boundary masking is completely removed, reflecting the periodic and dynamically connected nature of global zonal circulation, while masking at the north-south boundaries is retained to respect the lack of physical adjacency across the poles. This asymmetric shift-and-mask mechanism allows information to propagate seamlessly along longitudes while preventing non-physical feature mixing across latitudinal extremes. By embedding domain-specific physical constraints directly into the attention operation, the Searth Transformer enables more faithful representation of global atmospheric dynamics compared to generic vision transformers.

*C. Relay Autoregressive fine-tuning*

To address the critical challenge of cumulative error growth in medium-range forecasting, we introduce a novel training paradigm named Relay Autoregressive (RAR) fine-tuning. While standard pre-training facilitates the capture of short-term atmospheric states, it often fails to model the continuous evolution of complex weather systems, leading to rapid error accumulation. To suppress the rapid growth of accumulated errors, the pretrained model have to be fine-tuned in an autoregressive manner. However, standard autoregressive fine-tuning methods are constrained by GPU memory, preventing a single model from learning long continuous weather evolution over extended periods (e.g., 10 days). The core of our proposed RAR fine-tuning approach is to bridge this gap by decomposing a long-term forecasting horizon (e.g., 15 days) into a sequence of computationally manageable sub-stage. As illustrated in **Fig. 2**, a long-term weather process is decomposed into multiple sub-stages. Within each stage $s \in 1, \ldots, M$, the model performs a k-step autoregressive rollout. The cumulative loss over these $k$ steps is calculated as $\mathcal{L}_s = \sum_{t=1}^{k} loss(\hat{X}_t, X_t)$, followed by a standard backpropagation to update the model parameters.

Then, gradient detachment mechanism is strategically applied at the boundary of each stage. Specifically, the final prediction of the $s$-th stage, $\hat{X}_{i+k}$, is detached from the current computational graph and serves as the input for the $(s + 1)$-th stage. This "relay" mechanism enables the model to inherit the cumulative errors and physical biases from preceding steps—forcing it to learn error-correction capabilities—while maintaining a constant peak memory footprint proportional only to $k$. Consequently, our YanTian model achieves a comprehensive understanding of 15-day weather development with a modest memory requirement (< 25 GB).

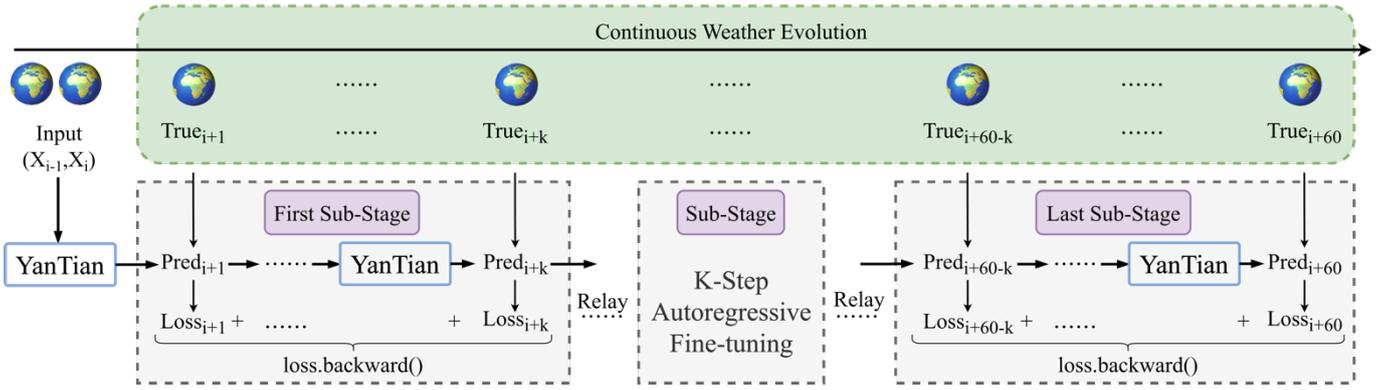

**Fig. 2.** Illustration of the Relay Autoregressive (RAR) fine-tuning strategy. The continuous weather evolution over a long forecast window is decomposed into multiple sub-stages, where the YanTian model performs k-step autoregressive rollouts within each sub-stage and accumulates the corresponding losses for backpropagation. At sub-stage boundaries, the final prediction is detached and relayed as the initial input for the next sub-stage, enabling effective error correction across stages while maintaining constant memory usage.

## IV. Experiments and Analysis

### A. Data Preparation

The YanTian model is trained using global atmospheric data at a spatial resolution of 1° (180×360 latitude-longitude grid) and a temporal resolution of 6 hours. The 1° data are derived from the 0.25° (721×1440 latitude-longitude grid) ERA5 global reanalysis by averaging adjacent grid points. ERA5 is the fifth-generation reanalysis product of the European Centre for Medium-Range Weather Forecasts (ECMWF) and is widely regarded as one of the most comprehensive and accurate atmospheric reanalysis datasets available [20]. Given that the native spatial resolution of ERA5 is 0.25°, the original fields are partitioned using a $4 \times 4$ window, and the 16 grid points within each window are averaged to obtain the corresponding 1° value. Because the number of ERA5 grid points in the meridional direction is 721, which is not divisible by 4, the Antarctic grid point is excluded during the averaging procedure. As a result, the horizontal spatial dimensions of the data are reduced from (721,1440) to (180,360).

The input and output of the YanTian model consist of the same set of variables, totaling 69 atmospheric variables, which are categorized into upper-air variables and surface variables. The upper-air variables include geopotential (Z), temperature (T), u component of wind (U), v component of wind (V), and relative humidity (R). Each upper-air variable has 13 pressure levels ranging from 50 hPa to 1000 hPa, specifically at 50, 100, 150, 200, 250, 300, 400, 500, 600, 700, 850, 925, and 1000 hPa. The surface variables comprise four quantities: mean sea level pressure (MSL), 2-m air temperature (T2M), 10-m u component of wind (U10), and 10-m v component of wind (V10).

A total of 30 years of data from 1987 to 2016 are employed for model training, with four times (00, 06, 12, and 18 UTC) selected each day. Data from 2017 to 2019 are used as the validation set, while data from 2020 are used exclusively for testing.

### B. Model Training Details

This section provides a detailed description of the training configuration of the YanTian model. The training procedure consists of two successive stages, namely pre-training and fine-tuning, similar to the approachs used for training FuXi [9]. The pre-training stage aims to endow the model with single-step forecasting capability, whereby meteorological fields from the two preceding time steps are used to predict the fields at the subsequent time step. The fine-tuning stage aims to enable the model to learn continuous weather evolution over a 15-day period, thereby mitigating the rapid accumulation of forecast errors. All training experiments are conducted on eight Nvidia A800 GPUS using the PyTorch framework [29]. To reduce GPU memory consumption, mixed-precision training with FP32 and FP16 arithmetic is adopted in conjunction with the gradient checkpointing [30], [31], such that the peak GPU memory usage during training is maintained below 25 GB. In both the pre-training and fine-tuning stages, a latitude-weighted mean absolute error (MAE) is employed as the loss function, which is defined as

$$Loss = \frac{\sum_{t_0=1}^{B} \sum_{\tau=1}^{T} \sum_{c=1}^{C} \sum_{i=1}^{H} \sum_{j=1}^{W} L_i \left| \hat{Y}_{c,i,j}^{t_0+\tau} - Y_{c,i,j}^{t_0+\tau} \right|}{B \times T \times C \times H \times W}. \tag{1}$$

Here, $B$, $T$, $C$, $H$, and $W$ denote the batch size, the number of autoregressive forecasting steps, the number of variables (channels), and the numbers of grid points in the latitudinal and longitudinal directions, respectively. The variable $t_0$ represents the forecast initialization date–time for each sample in a training batch, while $\tau$ denotes the lead time corresponding to the $T$ autoregressive steps. Specifically, $T = 1$ corresponds the pre-training stage, whereas $T \geq 2$ represents the fine-tuning stage. The indices $c$, $i$, and $j$ correspond to the variable, latitude, and longitude, respectively. $\hat{Y}_{c,i,j}^{t_0+\tau}$ and $Y_{c,i,j}^{t_0+\tau}$ denote the predicted value and the corresponding ground truth at spatial location $(i,j)$ for variable $c$ at time $t_0 + \tau$. Here, $L(i) = N_{\text{lat}} \times \frac{\cos \phi_i}{\sum_{i'=1}^{N_{\text{lat}}} \cos \phi_{i'}}$ denotes the weight at latitude $\phi_i$, where $N_{\text{lat}}$ represents the total number of grid points in the latitudinal direction.

During the pre-training stage, 30 years of data from 1987 to 2016 are used for training, and the model is trained for a total of $1 \times 10^5$ parameter update iterations. The AdamW optimizer is employed with $\beta_1 = 0.9$, $\beta_2 = 0.95$, and a weight decay coefficient of 0.1 [32], [33]. The initial learning rate is set to $2.5 \times 10^{-4}$ and is annealed to $1 \times 10^{-7}$ over $1 \times 10^5$ iterations using a cosine decay schedule [34]. The batch size is set to 4 per GPU, resulting in an effective batch size of 32 across eight GPUs. Notably, when the per-GPU batch size is set to 1, the memory consumption during the pretraining stage can be controlled within 10 GB. Scheduled DropPath with a dropping ratio of 0.2 is applied to alleviate overfitting [35]. The entire pre-training stage requires approximately three days to complete.

The fine-tuning stage is conducted to further enhance the model's capability for modeling continuous weather evolution. The Relay Autoregressive fine-tuning strategy is proposed, in which a long continuous weather process is decomposed into a sequence of temporally contiguous sub-stages. In this study, the model is trained to learn continuous weather evolution over a 15-day period. The 15-day sequence is divided into 15 contiguous sub-stages, each with a duration of 1 day. Consequently, a complete 15-day training sample consists of 15 successive four-step autoregressive fine-tuning processes. During fine-tuning, the batch size is set to 1 per GPU. Each GPU is used to train on 1000 independent 15-day continuous weather sequences, resulting in a total of 8000 long-range weather evolution samples across eight GPUs. The entire fine-tuning process is completed within approximately 8 h, using a constant learning rate of $1 \times 10^{-7}$.

*C. Evaluation Metrics*

We follow [9] to evaluate the performance of the YanTian model using the latitude-weighted root mean square error (RMSE) and the anomaly correlation coefficient (ACC), which are defined as follows:

$$RMSE(c,\tau) = \frac{1}{|D|} \sum_{t_0 \in D} \sqrt{\frac{1}{H \times W} \sum_{i=1}^{H} \sum_{j=1}^{W} L_i \left( \hat{Y}_{c,i,j}^{t_0+\tau} - Y_{c,i,j}^{t_0+\tau} \right)^2} \tag{2}$$

$$ACC(c,\tau) = \frac{1}{|D|} \sum_{t_0 \in D} \frac{\sum_{i,j} L_i \hat{A}_{c,i,j}^{t_0+\tau} A_{c,i,j}^{t_0+\tau}}{\sqrt{\sum_{i,j} L_i \left( \hat{A}_{c,i,j}^{t_0+\tau} \right)^2 \sum_{i,j} L_i \left( A_{c,i,j}^{t_0+\tau} \right)^2}} \tag{3}$$

In these formulations, $RMSE(c,\tau)$ denotes the average RMSE of variable $c$ at lead time $\tau$ computed over the entire test dataset, while $ACC(c,\tau)$ represents the corresponding average ACC. The set $D$ denotes the test dataset. The anomaly field $A$ is defined as the difference between the ground-truth field $Y$ and the corresponding climatological mean, whereas $\hat{A}$ denotes the difference between the predicted field $\hat{Y}$ and the climatological mean. The climatological mean is computed from the ERA5 reanalysis data over the period from 1993 to 2019. We also use the normalized RMSE difference between model A and baseline B calculated as $(RMSE_A - RMSE_B)/RMSE_B$, and the normalized ACC difference represented by $(ACC_A - ACC_B)/(1 - ACC_B)$. Negative values in normalized RMSE difference and positive values in normalized ACC difference indicate that model A performs better than the baseline model B.

*D. Experimental Results*

In this study, data from 2020 are used to evaluate the performance of YanTian. Forecasts are initialized twice daily at 00:00 UTC and 12:00 UTC, producing predictions with lead times of up to 10 days. In parallel, the corresponding 0.25° resolution forecasts from PanGu, GraphCast, FuXi, and the ECMWF HRES are downloaded from WeatherBench 2 for the same initialization times and forecast horizons [36]. All forecasts are processed to 1° resolution using window-averaging procedure and are compared against the YanTian model. It should be noted that, for the evaluation of AI-based models, ERA5 reanalysis data are used as the verification reference because these models are trained on reanalysis fields. For the evaluation of HRES, following the assessment protocols adopted in GraphCast [16] and FuXi [9], HRES-fc0 is used as the reference, where HRES-fc0 corresponds to the first forecast time step of each HRES forecast cycle.

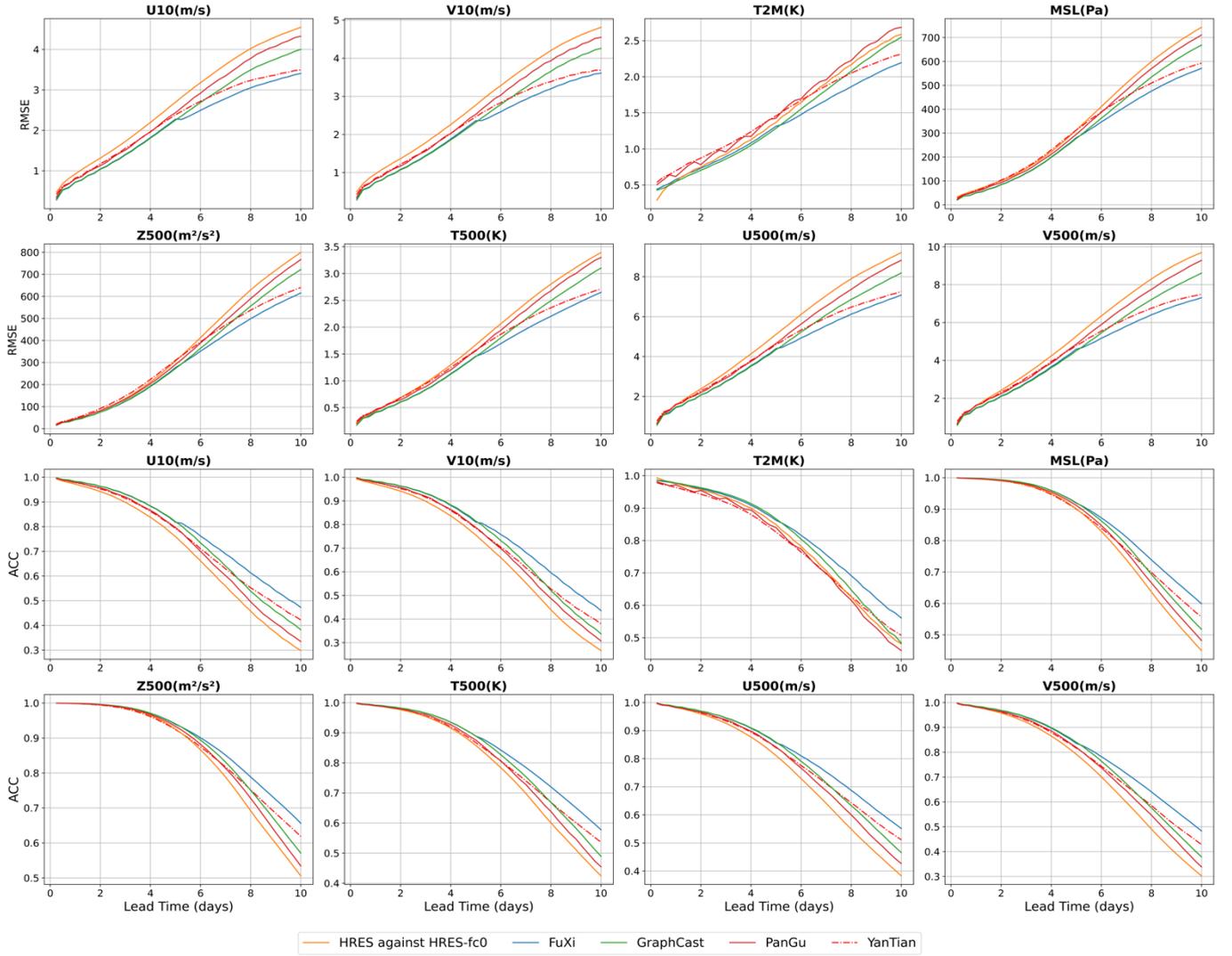

**Fig. 3.** Comparison of the globally averaged latitude-weighted RMSE (first and second rows) and ACC (third and fourth rows) of the HRES, FuXi, GraphCast PanGu, and YanTian for 4 surface variables, such as U10, V10, T2M and MSL, and 4 upper-air variables at the pressure level of 500 hPa, including Z500, T500, U500, and V500, using testing data from 2020. AI models are evaluated against the ERA5 reanalysis dataset, and ECMWF HRES is evaluated against HRES-fc0.

**Fig. 3** illustrates the time series of the globally averaged, latitude-weighted RMSE and ACC for ECMWF HRES, FuXi, GraphCast, PanGu, and YanTian, evaluated on four surface variables (U10, V10, T2M, and MSL) and four upper-air variables at the 500-hPa pressure level (Z500, T500, U500, and V500). The results shows forecasts up to 10 days, consistent with the evaluation range of most models in weatherbench2 [36], while 15-day forecasts are reported in the ablation study of the RAR fine-tuning. Overall, ECMWF HRES (evaluated against HRES-fc0) provides a strong baseline across different variables and lead times, while the AI-based models consistently outperform HRES. During the short-range forecast period (0–5 days), the performance of different forecasting systems is comparable, with GraphCast and FuXi exhibiting the highest skill. Owing to its relatively coarser spatial resolution, the YanTian model is constrained in its ability to capture fine-scale structures during the early forecast period; nevertheless, its overall performance remains close to that of PanGu. In the medium- to extended-range forecasts (approximately days 5–10), performance differences among models become increasingly pronounced, with YanTian and FuXi achieving the best results. Both models are able to maintain higher forecast skill over longer lead times, with a noticeably slower growth of accumulated errors, and their superior performance becomes increasingly significant as lead times increase. Using an ACC value of 0.6 as the threshold to measure a skillful weather forecast, we found that YanTian extends the skillful forecast lead time compared to ECMWF HRES, especially pushing the lead time of Z500 from 9 to 10.3 days. Overall, these evaluation results demonstrate the superior performance of the proposed Searth Transformer architecture and the RAR fine-tuning strategy introduced in this work.

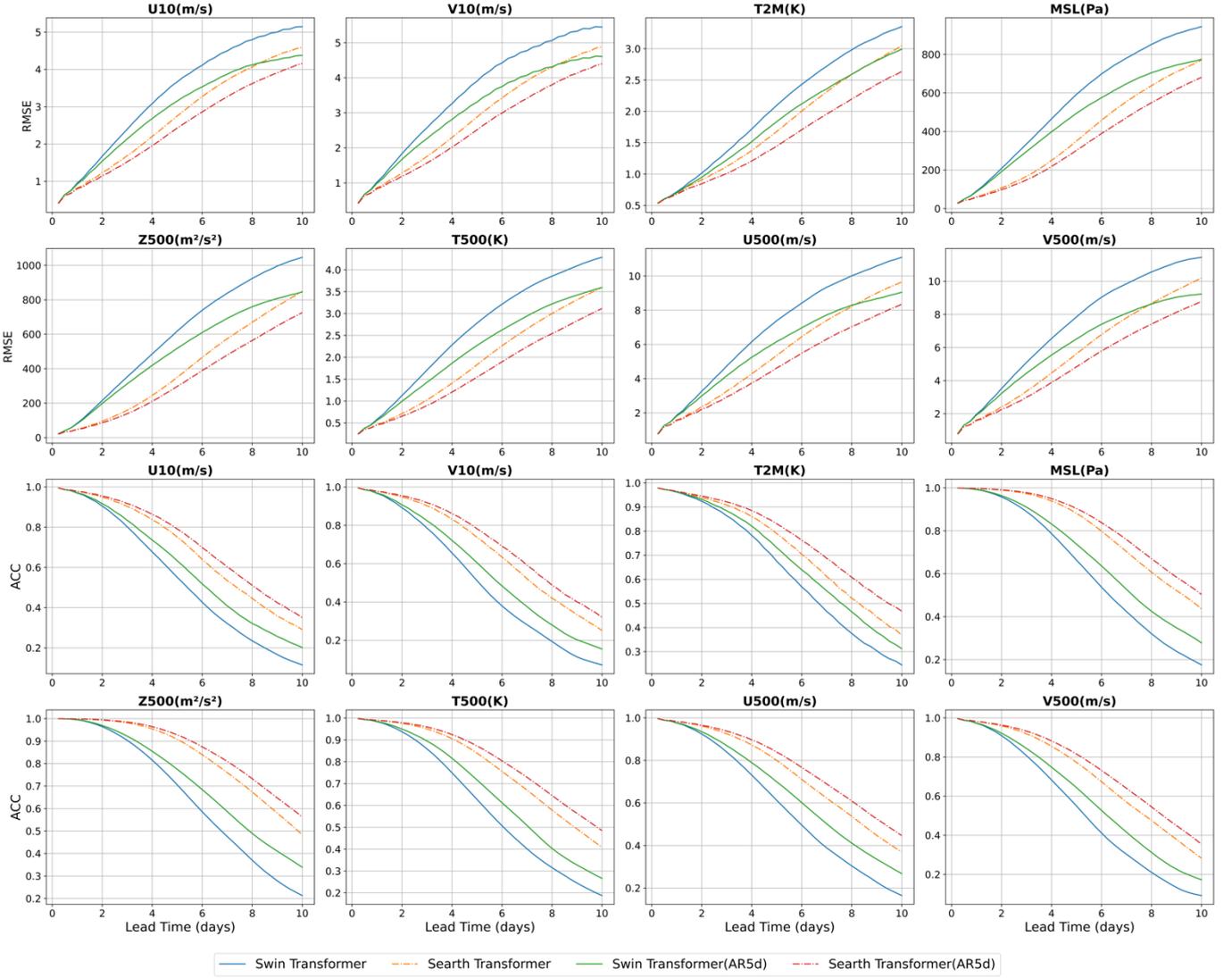

**Fig. 4.** Effects of the Searth Transformer. The first and second rows show the time series of globally averaged, latitude-weighted RMSE, while the third and fourth rows present the ACC. Each subfigure corresponds to one variable. Dashed lines denote models built with the Searth Transformer, whereas solid lines indicate models based on the Swin Transformer. AR5d represents models fine-tuned with 5-day autoregressive training, while the others correspond to pre-trained models.

*E. Ablation Study*

In this study, a systematic ablation study was conducted around the key designs of the proposed model, namely the Searth Transformer and the RAR fine-tuning strategy, to quantitatively evaluate the contributions of different algorithms and training strategies to overall performance. Specifically, the ablation analysis focuses on two aspects: first, the role of the Searth

Transformer in modeling and predicting the evolution of global atmospheric dynamics; and second, the effectiveness of the RAR fine-tuning strategy in enhancing the capability for continuous weather variation forecasting and mitigating error accumulation. By selectively replacing algorithms or modifying training hyperparameters while keeping the remaining network architecture and training configurations unchanged, prediction accuracy and error characteristics under different settings are systematically compared. This design aims to isolate the independent contributions of each component and to assess their synergistic benefits, thereby providing empirical evidence for the rationality and necessity of the overall effectiveness of the proposed model.

1) **Effect of the Searth Transformer**

In the ablation study of the Searth Transformer architecture, two groups of comparative experiments were designed, namely a pre-training group and a fine-tuning group:

a) *Pre-training group*: The Searth Transformer experiment employs the proposed Searth Transformer architecture and evaluates its forecasting performance after pre-training. In contrast, the Swin Transformer experiment replaces the Searth Transformer with the original Swin Transformer architecture while keeping all other components identical to the YanTian model, and conducts pre-training under the same configuration as YanTian..

b) *Fine-tuning group*: In the Swin Transformer (AR5d) experiment, the pretrained Swin Transformer model is further fine-tuned using a 20-step (5-day) autoregressive fine-tuning strategy and subsequently evaluated. Similarly, in the Searth Transformer (AR5d) experiment, the pretrained Searth Transformer model undergoes the same 20-step (5-day) autoregressive fine-tuning. This comparison aims to evaluate the performance improvements brought by the Searth Transformer after learning continuous atmospheric processes. To exclude the influence of RAR fine-tuning, the classical autoregressive fine-tuning strategy, which is used in GraphCast, is adopted in this group.

As showed in **Fig. 4**, across both the pre-training and fine-tuning groups, the Searth Transformer consistently achieves lower RMSE values and higher ACC scores than the corresponding Swin Transformer across all variables and forecast lead times, demonstrating its superior forecasting and representation capability. Notably, in cross-group comparisons, the Searth Transformer model trained with pre-training alone outperforms the fully trained Swin Transformer model (with both pre-training and fine-tuning) in terms of RMSE and ACC over the majority of lead times for all variables. This empirical evidence highlights the substantial performance gains obtained by embedding geophysical prior knowledge into the model design, and further underscores the strong capability of the Searth Transformer in modeling the global atmospheric circulation system.

2) **Effect of the RAR fine-tuning**

In the ablation study targeting RAR fine-tuning, five experiments were designed to systematically evaluate its effectiveness in improving the modeling of continuous weather processes. Specifically, the Baseline experiment adopts a pretrained Searth Transformer architecture and serves as the baseline for the RAR fine-tuning ablation study. The AR5d experiment applies the classical 20-step (5-day) autoregressive fine-tuning strategy to the pretrained model and acts as a control experiment for the 20-step (5-day) RAR fine-tuning setting, enabling a fair comparison between the two fine-tuning approaches after learning weather processes of identical temporal length. The RAR5d, RAR10d, and RAR15d experiments employ the proposed RAR fine-tuning strategy to train the model to learn continuous weather processes with durations of 5 days, 10 days, and 15 days, respectively, in order to assess the impact of learning weather processes over different temporal horizons on model performance.

From the results of **Fig. 5**, It can be observed that all models trained with continuous weather process learning exhibit superior cumulative error suppression and long-range forecasting capability compared with the Baseline, demonstrating the necessity of the fine-tuning stage. A comparison between AR5d and RAR5d indicates that the two approaches achieve comparable performance in terms of RMSE and ACC. Specifically, within a 5-day lead time, both methods yield nearly identical RMSE and ACC values. Beyond 5 days, RAR5d is slightly inferior to AR5d in RMSE, while the ACC remains comparable. However, when evaluated using normalized RMSE and ACC, the RAR approach shows a slight degradation compared with the AR approach. TABLE I summarizes the computational resources required by different fine-tuning strategies and the corresponding skillful forecast lead time for Z500. The comparison between the AR5d and RAR5d experiments shows that the AR5d setting requires approximately 200 hours of training time with a peak GPU memory consumption of about 80 GB, whereas the RAR5d experiment completes within approximately 3 hours with peak memory usage below 25 GB. If the product of training time and GPU memory consumption is used as an indicator of computational cost, the AR approach requires more than 200 times the computational resources of the RAR approach. Given such a substantial disparity in computational cost, a slight degradation in performance is considered acceptable. These results indicate that the proposed RAR approach can effectively learn the evolution of continuous weather processes and achieve performance comparable to that of the classical autoregressive method, while substantially reducing computational cost, thereby highlighting its efficiency and practical value.

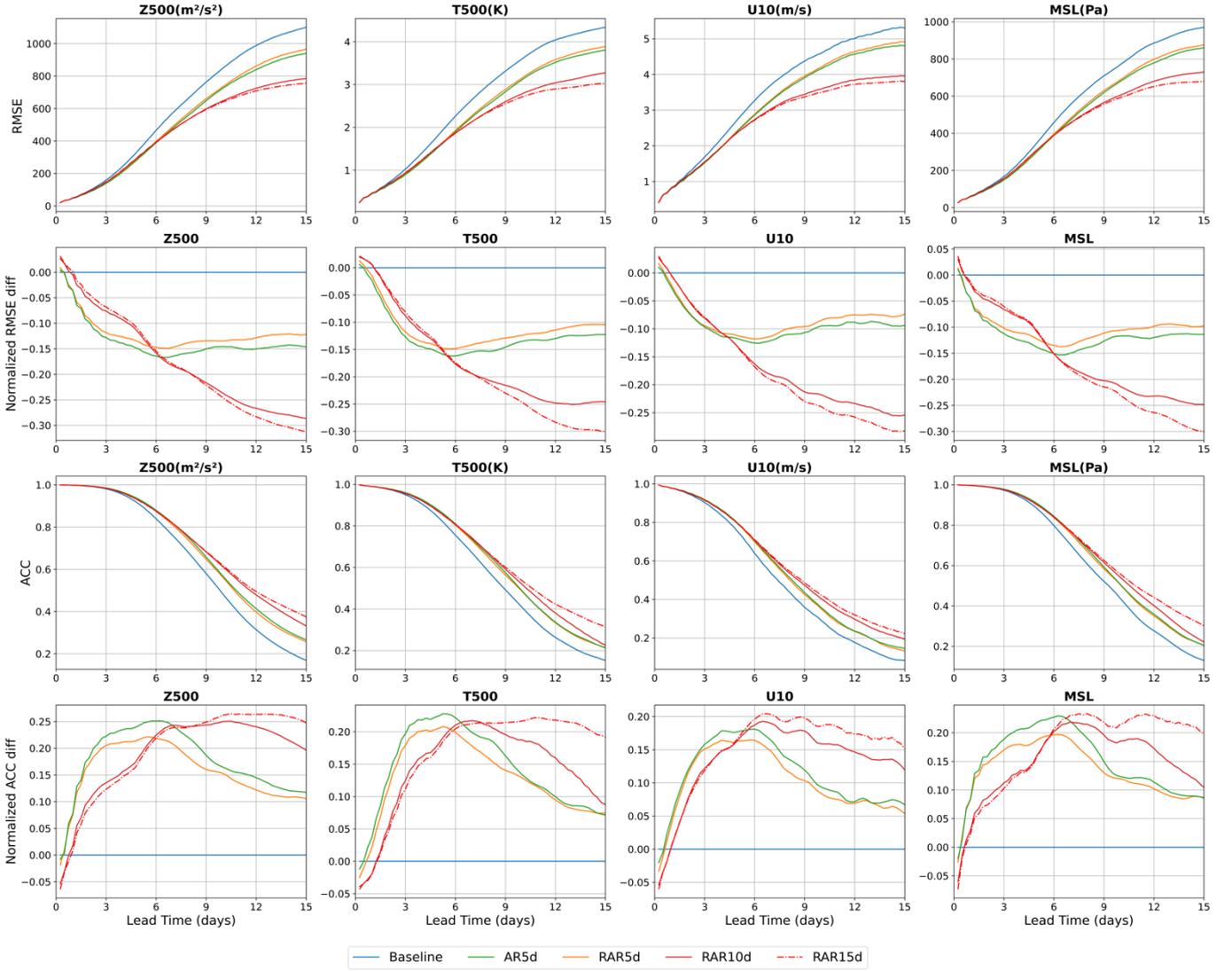

**Fig. 5.** Effects of the Relay Autoregressive fine-tuning strategy. From top to bottom, the four rows show absolute RMSE, normalized RMSE difference relative to the Baseline, anomaly correlation coefficient (ACC), and normalized ACC difference, respectively, for Z500, T500, U10, and MSL. The Baseline corresponds to the pretrained Searth Transformer model, whereas all other curves are obtained by applying different fine-tuning strategies on this pretrained model. AR5d denotes standard autoregressive fine-tuning for 5 days (20 steps), whereas RAR indicates Relay Autoregressive fine-tuning with learning horizons of 5 days (20 steps), 10 days (40 steps), and 15 days (60 steps) for RAR5d, RAR10d, and RAR15d respectively.

Furthermore, comparisons among the RAR5d, RAR10d, and RAR15d experiments reveal that, as the number of autoregressive steps used to learn continuous weather processes increases, the growth rate of RMSE with forecast lead time becomes slower, and the degradation of ACC is correspondingly mitigated. This effect is particularly pronounced in the RAR5d and RAR10d experiments. In terms of normalized RMSE and ACC, as the number of autoregressive steps increases, the RAR approach exhibits a slight degradation in performance at shorter lead times, but achieves a substantial improvement in forecasting skill at medium and longer lead times. The results indicates that learning longer continuous weather processes can significantly enhance the model's capability in medium- to long-range forecasting and more effectively suppress the accumulation of forecast errors.

TABLE I

COMPARISON OF GPU MEMORY USAGE, TRAINING TIME, AND Z500 FORECAST SKILL AMONG DIFFERENT FINE-TUNING STRATEGIES.

| Model | GPU Memory Usage | Training Time | ACC = 0.6 |
| --- | --- | --- | --- |
| **Baseline** | - | - | ~ 8.7d |
| **AR5d** | ~ 80GB | ~ 200h | ~ 9.7d |
| **RAR5d** | < 25GB | ~ 3h | ~ 9.6d |
| **RAR10d** | < 25GB | ~ 5h | ~ 10.2d |
| **RAR15d** | < 25GB | ~ 8h | ~10.3d |

## V. CONCLUSION

Accurate global medium-range weather forecasting remains a central challenge in Earth system science and operational meteorology [37], with growing reliance on data-driven Transformer-based models [13]. Physics-constrained artificial intelligence has emerged as a major development direction in recent years [38]. However, most existing approaches in global mid-range weather forecasting directly inherit vision-oriented architectures and training paradigms, thereby neglecting fundamental physical characteristics of the Earth system, such as zonal periodic continuity, meridional non-periodicity, and the intrinsic constraints of long-horizon autoregressive learning.

In this work, a physics-aware framework is proposed to address these structural and optimization limitations. The framework introduces two tightly coupled innovations. First, the Shifted Earth Transformer explicitly embeds Earth system topological priors into window-based self-attention through an asymmetric shift-and-mask mechanism, enabling physically consistent and computationally efficient global information exchange. This design substantially improves the representation of large-scale atmospheric circulation compared with directly adopting conventional Swin Transformer architectures [7]. Second, the Relay Autoregressive fine-tuning strategy decouples GPU memory consumption from forecast length by employing relay-style temporal unrolling with gradient detachment, allowing a single model to learn continuous atmospheric evolution over extended horizons while mitigating cumulative error growth. Experiments on ERA5 reanalysis data at 1° resolution demonstrate that the proposed approaches achieve superior forecast accuracy and stability, outperforming strong AI-based baselines in terms of latitude-weighted RMSE and ACC. Notably, these gains are achieved with modest computational resources, with peak memory usage maintained below 25GB. In addition, we apply bilinear interpolation to downscale the YanTian model outputs to 0.25° resolution as a simple baseline approach and compare the results with those of other models as showed in **Fig. 6** [39]. As direct bilinear interpolation neglects local information such as topographic effects [40], a minor reduction in forecast accuracy is observed for some variables (e.g., T2M). Nevertheless, the 0.25° results remain satisfactory. Overall, this study demonstrates that the framework composed of the Searth Transformer and the RAR fine-tuning strategy offers a viable and scalable pathway toward more accurate, stable, and resource-efficient AI-based global weather forecasting systems.

This study still has several limitations. First, the current approach only incorporates the zonal periodic continuity of the Earth system as a physical prior into the model design, and does not explicitly address the physical inconsistencies introduced when projecting the spherical Earth onto a latitude–longitude grid. In particular, the substantial enlargement of grid cell areas in high-latitude regions, including the polar areas, remains unresolved, which constrains the model's ability to accurately represent polar dynamics and the associated energy distribution. Second, the use of coarse-resolution reanalysis data at 1° objectively limits the upper bound of achievable predictive accuracy. Moreover, during the relay autoregressive fine-tuning stage, only 8000 continuous weather samples are used, which further restricts the model's capacity to fully capture long-term weather evolution processes and to effectively mitigate the growth of cumulative forecast errors.

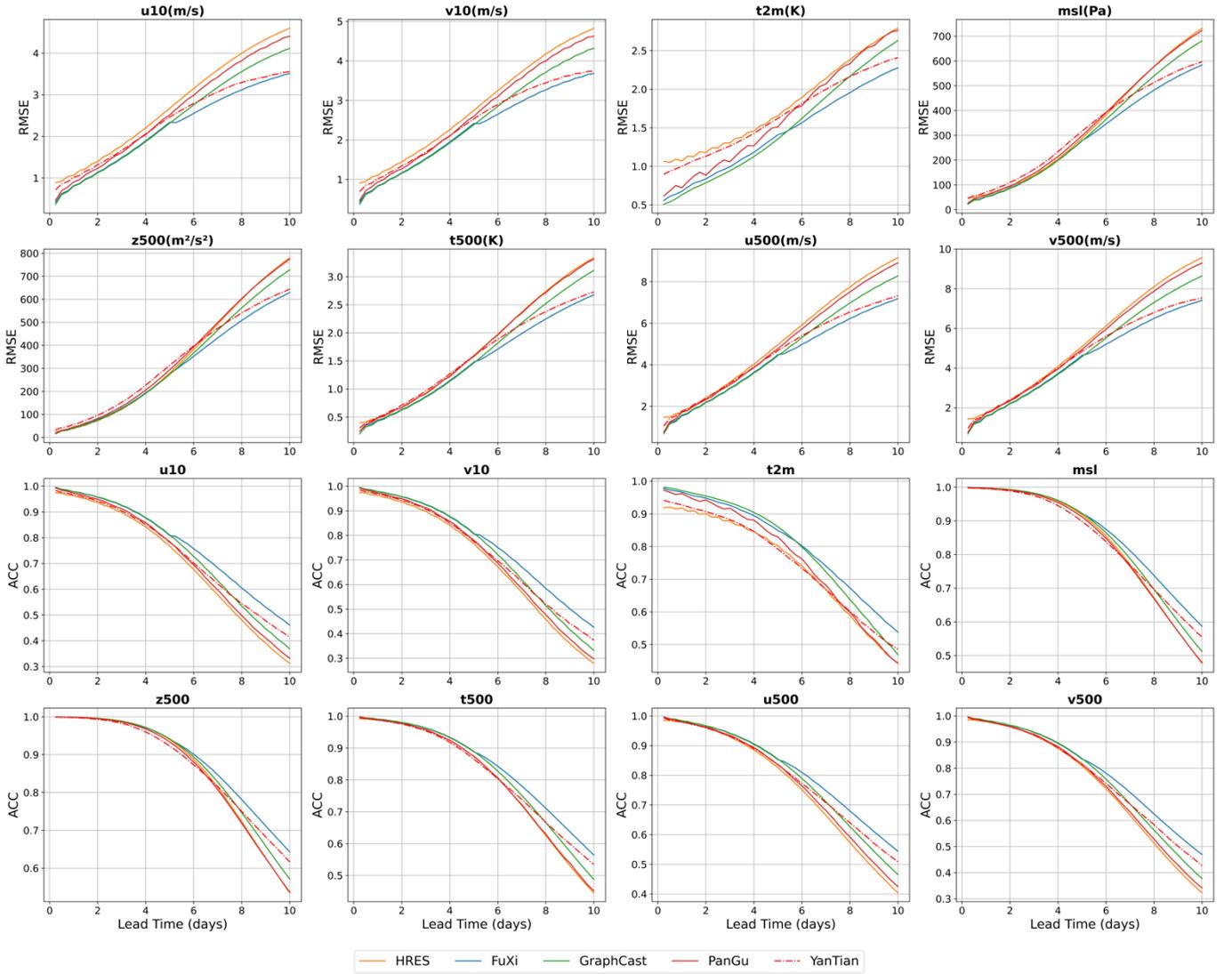

**Fig. 6.** Comparison of the globally averaged latitude-weighted RMSE (first and second rows) and ACC (third and fourth rows) of the HRES, FuXi, GraphCast PanGu, and YanTian for 4 surface variables, such as U10, V10, T2M and MSL, and 4 upper-air variables at the pressure level of 500 hPa, including Z500, T500, U500, and V500, using testing data from 2020. The results are evaluated at a 0.25° resolution. AI models are evaluated against the ERA5 reanalysis dataset, and ECMWF HRES is evaluated against HRES-fc0.

Looking ahead, future work will build upon the current framework by further incorporating the spherical geometry and dynamical characteristics of the Earth system, with the aim of embedding more comprehensive physical constraints into Transformer architectures. This is expected to continuously enhance the physical consistency and representational capability of the Transformer for the Earth system modeling. In addition, by introducing training data with higher spatial resolution and substantially strengthening the learning of continuous weather processes, we aim to develop a global medium-range weather forecasting foundation model that achieves a more favorable balance among accuracy, stability, and computational efficiency.

ACKNOWLEDGMENT

This work was funded by the National Natural Science Foundation of China (Grant No. 42275158) and the Guangdong Basic and Applied Basic Research Foundation (Grant 2025A1515010928). We thank for the technical support of the National Large Scientific and Technological Infrastructure "Earth System Numerical Simulation Facility".